\documentclass[10pt,twocolumn,letterpaper]{article}

\usepackage{cvpr}              % To produce the CAMERA-READY version

\usepackage{graphicx}
\usepackage{multirow}
\usepackage{booktabs}
\usepackage{algorithm}
\usepackage{algpseudocode}

\definecolor{cvprblue}{rgb}{0.21,0.49,0.74}
\usepackage[pagebackref,breaklinks,colorlinks,allcolors=cvprblue]{hyperref}

\def\confName{CVPR}
\def\confYear{2027}

\title{Hindsight Forcing: Learning Better History from the Future in Autoregressive Video Diffusion}

\author{
Haonan An$^1$ \quad Yihang Tao$^{1}$ \quad Yu Guo$^{1}$ \quad Renlong Chen$^2$\footnotemark[1] \quad
Jiasheng Zhang$^2$ \quad Yuguang Fang$^1$\footnotemark[1]\\
$^1$City University of Hong Kong, Hong Kong \quad
$^2$Philo AI, Singapore \\
{\tt\small 
(haonanan2-c, yihang.tommy, yu.guo)@my.cityu.edu.hk, }
{\tt\small 
(my.fang)@cityu.edu.hk,
} \\
{\tt\small 
(reo, alex)@philoai.com
}
}

\begin{document}
\maketitle
\begin{abstract}
Few-step autoregressive video diffusion enables real-time streaming generation, with each generated chunk serving as historical context for subsequent generation. 
To avoid the memory explosion of backpropagation through time, previous self-rollout methods typically truncate cross-chunk gradients or detach historical latents. As a result, earlier denoising computations receive no downstream feedback through the generated history on how to synthesize frames that facilitate subsequent generation, leading to visual drift and structural collapse as rollouts extend. To address this limitation, we propose Hindsight Forcing, which allows historical chunks to learn directly from future outcomes by decoupling forward generation from backward credit assignment via a latent straight-through bridge. In the forward pass, each chunk completes a full multi-step rollout to update the KV cache. In the backward pass, the bridge routes gradients from future losses back to a single selected denoising step per historical chunk, training both the context writer and historical denoiser on shared parameters without retaining the full trajectory graph. Remarkably, Hindsight Forcing requires no additional model parameters, architecture modifications, or separate trajectory replay. Extensive experiments on VBench demonstrate that Hindsight Forcing outperforms existing baselines in overall visual quality and dynamic motion, while ablation studies validate the benefit of supervising historical denoising. Code is available at \url{https://github.com/haonanAN309/Hindsight-Forcing}.
\end{abstract}    
\section{Introduction}
\label{sec:intro}
Recent advances in video generation have achieved remarkable visual fidelity using large-scale Diffusion Transformers (DiTs)~\cite{peebles2023scalable, wan2025wan, polyak2024movie, yang2025cogvideox}. Despite their success, these foundation models typically employ bidirectional attention across the full sequence, requiring all frames to be jointly denoised. Users therefore have to wait for the entire sequence to be generated before viewing the output, resulting in high initial latency that hinders deployment in interactive scenarios such as world simulation, live content creation, and embodied intelligence~\cite{parker2025genie, bruce2024genie, yin2025slow}. To alleviate this bottleneck, autoregressive video diffusion generates video sequentially with causal temporal attention~\cite{yin2025slow, zhu2026causal, jin2025pyramidal}. When paired with few-step distillation~\cite{salimans2022progressive, song2023consistency} such as Distribution Matching Distillation (DMD)~\cite{yin2024one, yin2024improved}, these models can synthesize chunks in as few as four steps, enabling real-time streaming via KV caching~\cite{yin2025slow, zhu2026causal, huang2025self}. In this streaming process, each generated chunk becomes part of the historical context for subsequent generation. Its content therefore affects both the quality of the current output and how well subsequent chunks can continue from it, making the generation of useful history essential for sustained long-horizon coherence~\cite{henschel2025streamingt2v,song2025history}.

However, training models to synthesize such useful history remains challenging. 
Conventional teacher forcing conditions on ground-truth video prefixes during training, whereas inference relies on self-generated history~\cite{jin2025pyramidal, huang2025self, lamb2016professor}. This discrepancy induces exposure bias, allowing generation errors to accumulate across chunks and undermine temporal coherence during autoregressive rollout~\cite{ross2011reduction,bengio2015scheduled, ruhe2024rolling}. To mitigate this mismatch, recent self-rollout methods, such as Self Forcing~\cite{huang2025self}, train the student directly on histories produced by its own sequential generation. Ideally, generation losses on downstream chunks should also propagate through the generated history to the earlier denoising and cache-writing computations, allowing them to learn how to support future continuation. In practice, however, full backpropagation through time (BPTT) requires retaining activations across both chunks and denoising steps, incurring substantial memory costs~\cite{zhuang2026self}. Consequently, previous self-rollout methods~\cite{huang2025self, zhu2026causal} typically truncate cross-chunk gradients by treating the historical KV cache as a detached rollout state. 

To address this limitation, recent approaches restore gradients through historical KV construction. For example, SGF~\cite{zhuang2026self} recomputes context representations in a second pass, while RAVEN~\cite{lu2026raven} processes historical latents and noisy denoising states in an interleaved sequence. 
SGF+~\cite{su2026sgf+} further assigns separate parameters to context writing and denoising to resolve their gradient conflicts. 
However, all three methods treat historical latents as detached inputs. While downstream generation losses can supervise how these latents are encoded into memory, their gradients stop at the latent inputs and cannot reach the earlier denoising computations that generated their content. As a result, the historical denoiser lacks direct feedback through the history-conditioning path on how to synthesize frames that support coherent future continuation. 

To extend future supervision from history encoding to history generation, we propose \textbf{Hindsight Forcing}, a framework that allows historical chunks to learn directly from future generation outcomes. Our key insight is to decouple the historical context used in forward generation from the gradient path used for backward credit assignment through a \textit{latent straight-through bridge}. Directly backpropagating through the full rollout incurs prohibitive memory costs, whereas constructing history from an intermediate denoising estimate exposes subsequent chunks to a different context distribution than the full-rollout outputs used at inference~\cite{huang2025self, zhuang2026self}. Our bridge therefore preserves each chunk's full-rollout output as
history while routing gradients to the clean estimate from a single selected denoising step through a surrogate gradient. Specifically, the training procedure operates as follows:
% Rather than backpropagating through every denoising step across chunks, we use a surrogate gradient to connect each chunk’s final denoised output to the clean estimate from a single selected denoising step. Specifically, the training procedure operates as follows:
\begin{itemize}
    \item \textbf{Forward pass}: Chunks are generated strictly sequentially. Each chunk completes a full multi-step denoising trajectory before updating the historical KV cache through a differentiable context writer, ensuring that downstream generation conditions on the final full rollout output rather than an intermediate clean estimate.
    \item \textbf{Backward pass}: Rather than retaining the computational graph across all denoising steps, the bridge routes gradients from downstream chunk losses back to a single selected denoising step per historical chunk. This allows future supervision to update both the context writer and the historical denoiser on shared parameters.
\end{itemize}
This formulation provides three key practical advantages over prior methods. First, it achieves effective cross-chunk credit assignment without introducing additional model parameters, maintaining a single set of shared weights for context writing and denoising instead of assigning separate parameters to the two roles~\cite{su2026sgf+}. Second, it builds gradient paths directly within the autoregressive rollout, requiring no separate trajectory replay or second parallel reconstruction pass as in SGF~\cite{zhuang2026self}. Finally, the bridge integrates into the standard sequential rollout without reorganizing historical latents and noisy denoising states into an interleaved training sequence as in RAVEN~\cite{lu2026raven}. Used only during training, it requires no architectural modifications and preserves standard multi-step autoregressive inference with KV caching and no additional computation.
\noindent Our main contributions are summarized as follows:
\begin{itemize}
\item We identify the detached-latent limitation in existing autoregressive video distillation methods, showing that stopping gradients at historical latents prevents future supervision from reaching the earlier denoising computations that generated them.
\item We propose Hindsight Forcing, which couples full multi-step forward rollouts with backward credit assignment via a latent straight-through bridge, enabling downstream losses to supervise both the context writer and the historical denoiser on shared parameters.
\item We conduct extensive experiments on
VBench~\cite{huang2024vbench}, demonstrating that our proposed Hindsight Forcing improves both video quality and temporal dynamics. Through component ablations and gradient analysis, we validate the benefit of extending future supervision beyond history encoding to the denoising computations that generate historical content, offering insight into the future design of autoregressive video diffusion frameworks.
\end{itemize}
\section{Related Work}
\label{sec:related_work}

% Required packages:
% \usepackage{booktabs}
% \usepackage{multirow}

\begin{table*}[t]
\centering
\small
\setlength{\tabcolsep}{7pt}
\renewcommand{\arraystretch}{1.15}
\caption{
Comparison of training-time history construction and cross-chunk credit assignment.
}
\label{tab:paradigm_comparison}
\begin{tabular}{@{}lccccc@{}}
\toprule
\multirow{2}{*}{\textbf{Method}}
& \multirow{2}{*}{\textbf{History Source}}
& \multicolumn{2}{c}{\textbf{Future Gradient To}}
& \multirow{2}{*}{\textbf{Gradient Mechanism}}
& \multirow{2}{*}{\textbf{Extra Params}} \\
\cmidrule(lr){3-4}
&
& \textbf{Encoding}
& \textbf{Denoising}
&
& \\
\midrule
CausVid~\cite{yin2025slow}          & Noisy data          & Yes$^{*}$   & N/A          & Joint forward    & 0 \\
Self Forcing~\cite{huang2025self}   & Selected-step       & No          & No           & Detached cache   & 0 \\
Causal Forcing~\cite{zhu2026causal} & Selected-step       & No          & No           & Detached cache   & 0 \\
RAVEN~\cite{lu2026raven}            & Final-step          & Yes         & No           & Interleaved pass & 0 \\
SGF~\cite{zhuang2026self}           & Selected-step       & Yes         & No           & 2-pass replay    & 0 \\
SGF+~\cite{su2026sgf+}             & Selected-step       & Yes         & No           & 2-pass replay    & $+100\%$ \\
\midrule
\textbf{Hindsight Forcing (Ours)}   & \textbf{Final-step} & \textbf{Yes} & \textbf{Yes} & \textbf{LST bridge} & \textbf{0} \\
\bottomrule
\end{tabular}

\vspace{3pt}
\begin{minipage}{0.98\textwidth}
\footnotesize
$^{*}$CausVid encodes data-derived noisy context without a student history-generation trajectory.
Other methods use self-generated clean estimates from a sampled step (\emph{Selected-step}) or the complete rollout (\emph{Final-step}).
Gradients denote cross-chunk feedback through history.
Extra parameters are relative to one student backbone; SGF+ uses full parameter separation.
\end{minipage}
\end{table*}

\subsection{Autoregressive Video Diffusion}
\label{sec:ar_video_diffusion}

Autoregressive video diffusion combines sequential generation with within-chunk iterative denoising to support streaming video synthesis~\cite{jin2025pyramidal, teng2025magi, chen2025skyreels}. During training, Diffusion Forcing~\cite{chen2024diffusion} trains causal sequence models with independent per-token noise levels, creating a training--inference mismatch when inference uses fully denoised history. To reduce denoising costs, CausVid~\cite{yin2025slow} distills bidirectional teachers into few-step causal generators using Distribution Matching Distillation (DMD). Building on this paradigm, Self Forcing~\cite{huang2025self} mitigates exposure bias through autoregressive self-rollouts, while Causal Forcing~\cite{zhu2026causal} addresses the teacher--student architectural gap through causal ODE distillation. Operating within this few-step self-rollout framework, we investigate how losses on subsequent chunks can provide feedback to the earlier denoising computations that generated their historical context.

\subsection{Cross-Chunk Credit Assignment}
\label{sec:cross_chunk_credit_assignment}

Recent approaches enable cross-chunk credit assignment by allowing downstream generation losses to supervise historical context encoding. For example, RAVEN~\cite{lu2026raven} processes historical latents and noisy states in an interleaved sequence, while SGF~\cite{zhuang2026self} reconstructs context representations in a second pass. Building on SGF, SGF+~\cite{su2026sgf+} separates context-writing and denoising parameters to address conflicting updates. In all three methods, however, historical latents remain detached, so this supervision cannot reach the earlier denoising computations that generated them. In contrast, our method bridges this gap by propagating supervision to the historical denoiser, requiring neither parameter duplication nor separate trajectory replay (see \Cref{tab:paradigm_comparison}).

\subsection{Straight-Through Estimation}
\label{sec:straight_through}

Our latent straight-through bridge is inspired by straight-through estimation, which preserves forward computations while using surrogate gradients during backpropagation~\cite{bengio2013estimating, yin2019understanding, huh2023straightening}. For example, VQ-VAE~\cite{van2017neural} applies this idea by retaining quantized representations in the forward pass but copying reconstruction gradients to encoder outputs in the backward pass. We apply this principle to autoregressive video diffusion: final denoised outputs provide historical context, while clean estimates from selected denoising steps receive gradients from subsequent generation, allowing downstream losses to supervise the historical denoiser without retaining the full denoising trajectory graph.
\section{Proposed Method: Hindsight Forcing}
\label{sec:method}

\subsection{Problem Setup and Gradient Truncation}
\label{sec:problem_setup}

Let a video be divided into $N$ temporal chunks, indexed by
$j \in \{0, \dots, N-1\}$, with a full rollout of each chunk
consisting of $K$ iterative denoising steps.
Let $G_\theta$ denote the denoiser with parameters $\theta$.
At denoising step $r \in \{1, \dots, K\}$ of chunk $j$,
it predicts a clean latent estimate:
\begin{equation}
\hat{x}_j^{(r)}
=
G_\theta(u_j^{(r)}, \sigma_r, c, H_{<j}),
\label{eq:denoising_prediction}
\end{equation}
where $u_j^{(r)}$ is the noisy latent at noise level $\sigma_r$,
$c$ denotes the conditioning information, including the text prompt
and an optional reference image, and $H_{<j}$ is the key-value
(KV) cache containing states from preceding chunks.

A full rollout produces the final clean latent
$x_j = \hat{x}_j^{(K)}$.
We denote the clean latent used to construct history by $h_j$:
full-rollout history uses $h_j = x_j$, whereas selected-step
history uses the clean estimate predicted at a randomly sampled
denoising step.
The context-writing call $W_\theta$ encodes $h_j$ into
$C_j = (K_j, V_j)$, the key and value tensors produced for chunk $j$ and cached for subsequent generation:
\begin{equation}
C_j
=
W_\theta(q_{\sigma_c}(h_j, \eta_j), \sigma_c, c, H_{<j}),
\label{eq:context_writing}
\end{equation}
where $q_{\sigma_c}$ denotes the forward noising operation at
context noise level $\sigma_c$, with
$\eta_j \sim \mathcal{N}(0,I)$.
The resulting KV states are appended to the historical cache,
$H_{<j+1} = H_{<j} \cup \{C_j\}$.
In our formulation, $W_\theta$ denotes a context-writing call
to the same network, sharing parameters $\theta$ with $G_\theta$.

This process establishes the following cross-chunk dependency:
\begin{equation}
\underbrace{u_j^{(1)} \!\to\! \dots \!\to\! h_j}_{\text{History generation}}
\xrightarrow{W_\theta}
\underbrace{C_j}_{\text{Encoding}}
\xrightarrow{\text{generation}}
\underbrace{\mathcal{L}_{>j}}_{\text{Future loss}}.
\label{eq:forward_dependency}
\end{equation}
Here, $\mathcal{L}_{>j}$ denotes the generation loss terms
associated with subsequent chunks $k > j$.
We focus on their gradients through the historical context.

Existing self-rollout methods truncate this feedback at
two distinct boundaries.
Below, $\operatorname{sg}$ means the stop-gradient operator,
which preserves its input value in the forward pass and blocks
gradient propagation in the backward pass.
\begin{enumerate}
    \item \textbf{Detaching the KV cache.}
    Subsequent chunks read detached KV states
    $\bar{C}_j = \operatorname{sg}(C_j)$.
    Since $\partial \bar{C}_j / \partial C_j = 0$,
    downstream gradients cannot reach either the context-writing
    computation that produced $C_j$ or the denoising computations
    that generated $h_j$~\cite{huang2025self,zhu2026causal}.

    \item \textbf{Detaching historical latents.}
    The context writer receives
    $\bar{h}_j = \operatorname{sg}(h_j)$,
    while context writing remains differentiable with respect
    to the writer parameters.
    Downstream losses can therefore train history encoding,
    but $\partial \bar{h}_j / \partial h_j = 0$ prevents their
    gradients from reaching the denoising computations that
    generated $h_j$~\cite{lu2026raven,zhuang2026self,su2026sgf+}.
\end{enumerate}
\begin{figure*}[t]
    \centering
    \includegraphics[width=\textwidth]{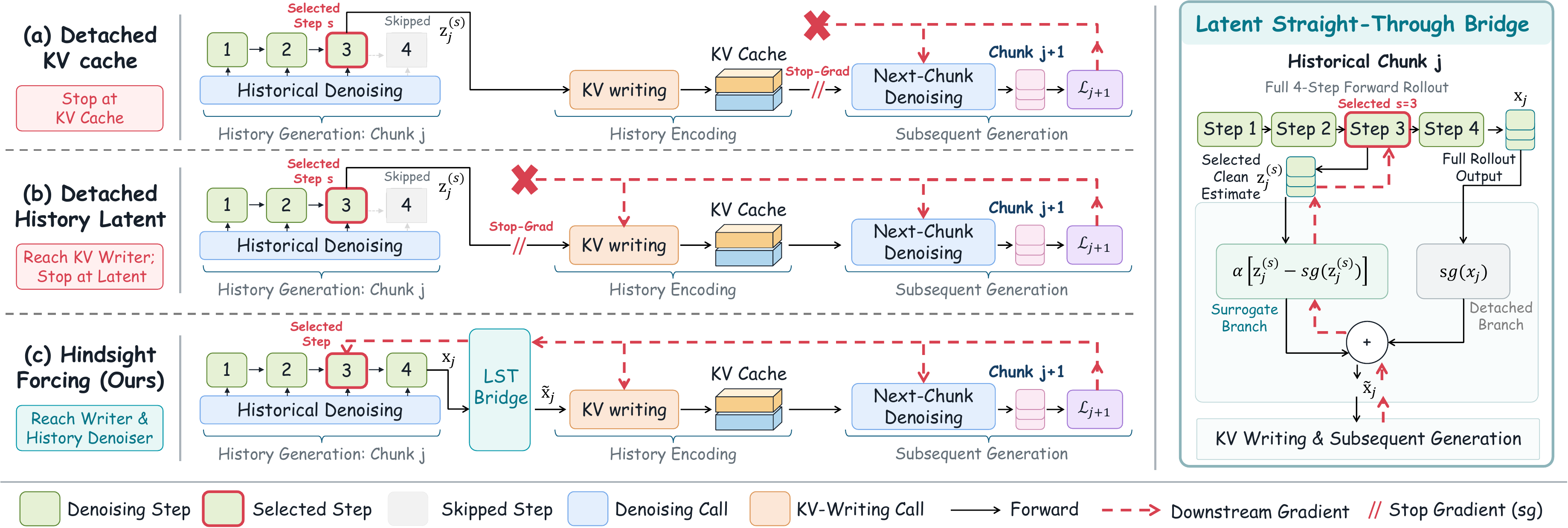}
    \caption{
    Overview of Hindsight Forcing.
    (a,b) In the illustrated training pipelines, downstream gradients stop
    at the historical KV cache or latent inputs, respectively.
    (c) Hindsight Forcing extends downstream supervision to both history
    encoding and generation.
    \textbf{Right:} The latent straight-through bridge preserves the full
    rollout output while routing gradients to a selected denoising step,
    without retaining the full trajectory graph.
    Teacher and critic models are omitted for brevity.
}
\label{fig:hf_workflow}
\end{figure*}
While full backpropagation through time across the rollout allows
downstream losses to supervise both history encoding and history
generation, it requires retaining activations across
all $N$ chunks and $K$ denoising steps per chunk, incurring substantial training memory costs.
This motivates a gradient path that reaches historical denoising
computations without retaining the full multi-step denoising graph.

\subsection{Latent Straight-Through Bridge}
\label{sec:lst_bridge}

To enable downstream supervision of historical denoising without
retaining the full multi-step denoising graph, we decouple the
historical content used in forward generation from the computation
used for backward credit assignment.
Each chunk completes the full $K$-step rollout before its output
$x_j$ is used to construct history for subsequent chunks.
Meanwhile, the clean estimate
$z_j^{(s)} = \hat{x}_j^{(s)}$ from a randomly sampled denoising step $s \in \{1,\dots,K\}$ provides the differentiable connection
to historical denoising.
We retain a denoising computation graph only at the selected step
and apply stop-gradient to the noisy latent input at every step.

To combine full-rollout history with selected-step credit assignment,
we construct the latent straight-through bridge:
\begin{equation}
\tilde{x}_j
=
\operatorname{sg}(x_j)
+
\alpha\left[z_j^{(s)}-\operatorname{sg}(z_j^{(s)})\right],
\label{eq:lst_bridge}
\end{equation}
where $\alpha > 0$ is a fixed gradient scaling factor.
In the forward pass, the bracketed term evaluates to zero,
so $\tilde{x}_j = x_j$.
Subsequent chunks therefore receive history constructed from
the full-rollout output rather than the selected-step estimate.
In the backward pass, the detached branch
$\operatorname{sg}(x_j)$ blocks gradients, while the selected-step
branch yields
\begin{equation}
\frac{\partial \tilde{x}_j}{\partial z_j^{(s)}} = \alpha I,
\label{eq:lst_gradient}
\end{equation}
where $I$ is the identity matrix.
The bridge thus transfers downstream gradients evaluated at
the full-rollout history to the selected-step estimate in the
same clean latent space.
This provides a surrogate backward path without differentiating
through the full multi-step denoising trajectory.

We set $h_j = \tilde{x}_j$ and construct the historical KV states
$C_j$ using \Cref{eq:context_writing}, retaining the context-writing
computation graph and its connection to the bridge.
As illustrated in \Cref{fig:hf_workflow}(c), downstream losses
$\mathcal{L}_{>j}$ can then propagate through $C_j$ to both the
context writer and the selected historical denoising step.
Let $\delta_{C_j} = \nabla_{C_j}\mathcal{L}_{>j}$ denote
the gradient arriving at these KV states.
The contribution to the shared parameter gradient from chunk
$j$'s context-writing and selected denoising calls is
\begin{equation}
\Delta_\theta^{(j)}
=
\underbrace{J_{W,\theta}^{\top}\delta_{C_j}}_{\text{Context writing}}
+
\underbrace{
\alpha J_{G,\theta}^{\top}J_{W,h}^{\top}\delta_{C_j}
}_{\text{Historical denoising}}.
\label{eq:dual_update}
\end{equation}
Here, $J_{W,\theta} = \partial C_j/\partial\theta$ and
$J_{G,\theta} = \partial z_j^{(s)}/\partial\theta$
are the local parameter Jacobians of the two calls,
with their inputs held fixed.
$J_{W,h} = \partial C_j/\partial h_j$ captures the dependence
of context writing on its historical input.

The first term trains the context writer to encode history
into KV representations useful for subsequent generation.
The second term passes through the bridge and trains the
historical denoiser to generate content that facilitates
future continuation.
Both roles are jointly optimized on the same parameters $\theta$,
without introducing auxiliary losses or additional model parameters.

\begin{figure*}[!t]
    \centering
    \includegraphics[width=\textwidth]{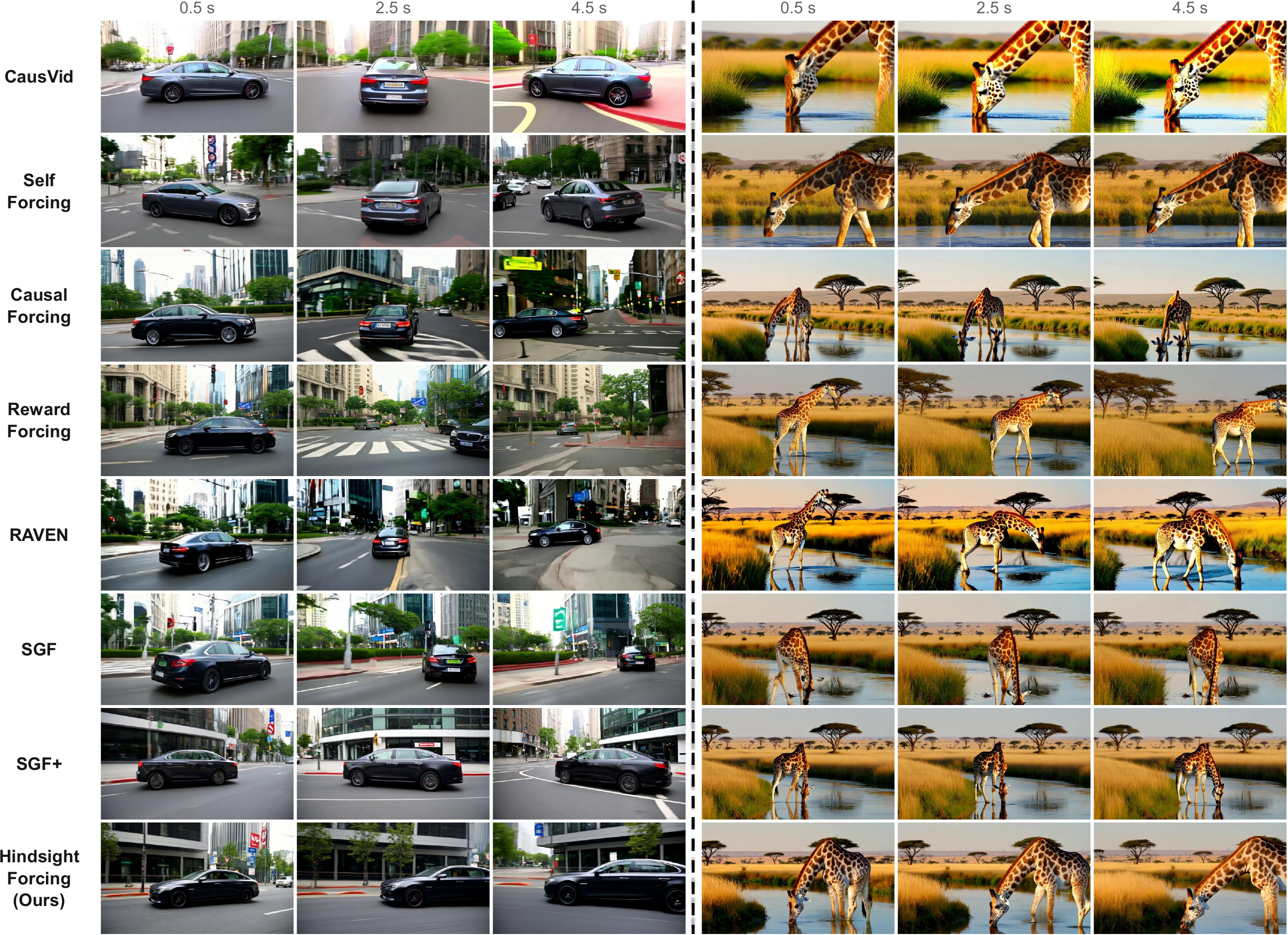}
    \caption{
       Qualitative comparison with existing methods.
    Frames sampled at 0.5, 2.5, and 4.5 seconds illustrate
    differences in motion plausibility, appearance consistency,
    and structural coherence.
    Additional short and long videos are
provided in the supplementary material.
    }
    \label{fig:qualitative}
\end{figure*}

\subsection{Training Objective and Computational Cost}
\label{sec:training_objective}

We train the generator $G_\theta$ as a student using Distribution
Matching Distillation (DMD)~\cite{yin2024one, yin2024improved}.
A frozen bidirectional teacher $D_T$ provides the target distribution,
while a diffusion critic $D_\phi$ with trainable parameters $\phi$
models the generator's current output distribution.
DMD supervision is applied to the concatenated selected-step
clean estimates $Z = [z_0^{(s)}, \dots, z_{N-1}^{(s)}]$,
while the full-rollout outputs $x_j$ provide historical context
through the bridge.
At a sampled noise level $\sigma$, we form
$Y_\sigma = (1-\sigma)\operatorname{sg}(Z) + \sigma\epsilon$,
where $\epsilon \sim \mathcal{N}(0,I)$.
The teacher and critic predict clean latents
$\mu_{\mathrm{real}} = D_T(Y_\sigma,\sigma,c)$ and
$\mu_{\mathrm{fake}} = D_\phi(Y_\sigma,\sigma,c)$, respectively.
Following the baseline DMD setup~\cite{zhu2026causal}, we form
the normalized update direction
$g = \operatorname{sg}[
\operatorname{Normalize}(\mu_{\mathrm{fake}}-\mu_{\mathrm{real}})
]$.
We apply this direction to the generator through the surrogate loss
\begin{equation}
\mathcal{L}_{\mathrm{student}}
=
\frac{1}{2D}
\left\|Z-\operatorname{sg}(Z-g)\right\|_2^2,
\label{eq:surrogate_loss}
\end{equation}
where $D$ is the number of scalar elements in $Z$.
The direct gradient with respect to the concatenated output is
$\nabla_Z\mathcal{L}_{\mathrm{student}} = g/D$.
Writing this loss as a sum of chunk-wise terms
$\mathcal{L}_{\mathrm{student}} = \sum_j \ell_j$,
the downstream objective introduced in
\Cref{sec:problem_setup} is
$\mathcal{L}_{>j} = \sum_{k>j}\ell_k$.

Through the bridge, the total gradient at $z_j^{(s)}$ combines
its direct DMD supervision with feedback from later chunks:
\begin{equation}
\frac{\mathrm{d}\mathcal{L}_{\mathrm{student}}}
{\mathrm{d}z_j^{(s)}}
=
\frac{g_j}{D}
+
\sum_{k>j} J_{k\leftarrow j}^{\top}\frac{g_k}{D},
\label{eq:total_gradient_accumulation}
\end{equation}
where $g_j$ is the chunk-$j$ slice of the video-level DMD
update direction $g$.
$J_{k\leftarrow j}$ is the surrogate Jacobian along the retained
history-conditioning paths
$z_j^{(s)} \to \tilde{x}_j \to C_j \to \cdots \to z_k^{(s)}$,
including dependencies through intervening chunks.
The second term, mapped through $J_{G,\theta}^{\top}$,
yields the historical-denoising contribution
in \Cref{eq:dual_update}.
During generator updates, the teacher and critic receive no
gradients from $\mathcal{L}_{\mathrm{student}}$.
The critic is trained separately using the baseline denoising
objective on detached generator samples $\operatorname{sg}(Z)$,
with the generator held fixed.

\begin{table*}[!ht]
\centering
\caption{
Quantitative comparison on VBench~\cite{huang2024vbench} and efficiency.
We evaluate overall performance and specific visual dimensions,
and report additional student parameters, peak training memory,
and inference time.
All VBench scores are scaled by 100.
For evaluation scores, bold and underline indicate the best
and second-best results, respectively.
}
\label{tab:main_results}
\vspace{-2pt}
\resizebox{\textwidth}{!}{%
\begin{tabular}{@{}l ccc ccc ccc@{}}
\toprule
\toprule
\multirow{2}{*}{\textbf{Method}}
& \multicolumn{3}{c}{\textbf{Overall Performance}}
& \multicolumn{3}{c}{\textbf{Specific Dimensions}}
& \multicolumn{3}{c}{\textbf{Efficiency}} \\
\cmidrule(lr){2-4} \cmidrule(lr){5-7} \cmidrule(lr){8-10}
& \shortstack{\textbf{VBench}\\\textbf{Total} $\uparrow$}
& \shortstack{\textbf{Quality}\\\textbf{Score} $\uparrow$}
& \shortstack{\textbf{Semantic}\\\textbf{Score} $\uparrow$}
& \shortstack{\textbf{Dynamic}\\\textbf{Degree} $\uparrow$}
& \shortstack{\textbf{Overall}\\\textbf{Consistency} $\uparrow$}
& \shortstack{\textbf{Temporal}\\\textbf{Flickering} $\uparrow$}
& \shortstack{\textbf{Extra Student}\\\textbf{Params} $\downarrow$}
& \shortstack{\textbf{Training Peak}\\\textbf{GPU Memory (GiB)} $\downarrow$}
& \shortstack{\textbf{Inference Time}\\\textbf{(s/video)} $\downarrow$} \\
\midrule
CausVid~\cite{yin2025slow}
& 83.40 & 84.69 & 78.26
& 70.00 & 26.19 & 98.86
& 0 & 24.21 & 5.425 \\

Self Forcing~\cite{huang2025self}
& 84.29 & 85.12 & 80.99
& 67.22 & 26.84 & \underline{99.05}
& 0 & 30.70 & 5.612 \\

Causal Forcing~\cite{zhu2026causal}
& 84.08 & 84.81 & \textbf{81.15}
& 84.17 & 26.64 & 98.05
& 0 & 30.70 & 5.603 \\

Reward Forcing~\cite{lu2026reward}
& \underline{85.07} & \underline{86.28} & 80.21
& 87.78 & 26.90 & --$^{*}$
& 0 & 41.38 & 5.376 \\

RAVEN~\cite{lu2026raven}
& 83.86 & 84.71 & 80.50
& \underline{91.67} & \textbf{26.94} & 98.01
& 0 & 30.27 & 5.631 \\

SGF~\cite{zhuang2026self}
& 84.70 & 85.66 & 80.87
& 74.72 & 26.79 & 98.51
& 0 & 36.80 & 4.996 \\

SGF+~\cite{su2026sgf+}
& 83.75 & 84.48 & 80.83
& 55.00 & 26.79 & 98.11
& +1.419B (+100\%) & 40.31 & 5.023 \\

\midrule
\textbf{Hindsight Forcing (Ours)}
& \textbf{85.74} & \textbf{86.89} & \underline{81.11}
& \textbf{92.22} & \underline{26.92} & \textbf{99.25}
& 0 & 38.08 & 5.619 \\
\bottomrule
\bottomrule
\end{tabular}%
}

\vspace{3pt}
\begin{minipage}{\textwidth}
\footnotesize
$^{*}$Reward Forcing's Temporal Flickering score is omitted
because only one video passed the static filter.
\end{minipage}

\end{table*}

\noindent\textbf{Computational Cost and Inference Invariance.}
The generator rollout executes $K$ denoising calls and one
context-writing call per chunk, while retaining graphs only
for the selected denoising calls and context writing.
The number of retained denoising graphs is therefore
$\mathcal{O}(N)$, compared with $\mathcal{O}(NK)$ for direct
backpropagation through every denoising step.
The gradient paths are established within the rollout,
requiring no separate trajectory replay or parallel reconstruction
pass as used in SGF and SGF+~\cite{zhuang2026self,su2026sgf+}.
At inference, the bridge is inactive and the model follows
the standard $K$-step autoregressive rollout with KV caching,
introducing no additional inference computation.

% Required packages in preamble:

\section{Experimental Results}
\label{sec:exp}

\subsection{Experimental Setup}
\label{sec:experimental_setup}

\noindent\textbf{Implementation Details.}
We adopt Wan2.1-T2V-1.3B~\cite{wan2025wan} as the generator, initialized from the causal ODE checkpoint of Causal Forcing~\cite{zhu2026causal}.
A frozen Wan2.1-T2V-14B serves as the bidirectional teacher,
while a trainable Wan2.1-T2V-1.3B serves as the diffusion critic.
We train on approximately 248K filtered and extended text
prompts from VidProM~\cite{wang2024vidprom}, with videos
generated online through autoregressive self-rollouts.
Training sequences contain seven chunks, with three latent frames
and four denoising steps per chunk.
We train on NVIDIA H100 80\,GB GPUs with a global batch size of 64,
using five critic updates per generator update.
The bridge gradient scale is set to $\alpha=1$ by default.
Our main results use the exponential moving average (EMA)
checkpoint after 150 generator updates.

\noindent\textbf{Evaluation Baselines.}
We compare Hindsight Forcing with CausVid~\cite{yin2025slow}, Self Forcing~\cite{huang2025self}, Causal Forcing~\cite{zhu2026causal}, Reward Forcing~\cite{lu2026reward}, RAVEN~\cite{lu2026raven}, SGF~\cite{zhuang2026self}, and SGF+~\cite{su2026sgf+}.
We use official checkpoints for CausVid, Self Forcing, and Causal Forcing.
Our method and ablations follow the official Causal Forcing codebase and setup, matching the training prompts, batch size, and budget of 150 generator updates used for its released checkpoint.
For the remaining baselines, we preserve their method-specific designs and evaluate EMA checkpoints obtained using the same causal ODE initialization, training prompts, batch size, and generator update budget as our method.
For RAVEN, we evaluate distillation without CM-GRPO.
For SGF+, we use full parameter separation between context writing and denoising.

\noindent\textbf{Evaluation Benchmarks.}
We evaluate generated videos using VBench~\cite{huang2024vbench}.
We report the VBench Total Score, Quality Score, and Semantic Score,
together with Dynamic Degree, Overall Consistency, and Temporal Flickering
to examine motion and consistency.
For each method, we generate five videos for each of
944 unique VBench prompts.
For the 75 Temporal Flickering prompts, we generate
20 additional candidates per prompt and apply the official
static filter, retaining up to five accepted videos per prompt.
Generation uses expanded prompts, while scoring uses the
original benchmark prompts.
Videos contain 81 frames at a resolution of $832\times480$
and 16 FPS, corresponding to approximately five seconds.
All methods share the same base evaluation prompts and random seeds.

\subsection{Qualitative Results}
\label{sec:qualitative}

\Cref{fig:qualitative} compares two examples that reveal
differences in scene continuity, appearance, and structural coherence.
In the driving scene, Hindsight Forcing keeps the car on the
road while preserving its appearance across changing viewpoints.
In contrast, CausVid, Causal Forcing, RAVEN, and SGF
steer the car onto the sidewalk, producing implausible trajectories.
SGF+ generates a distorted car body, while the car disappears from view in Reward Forcing.
In the giraffe scene, CausVid and RAVEN exhibit noticeable
color drift, while Self Forcing and Reward Forcing produce
little visible motion.
Causal Forcing generates a duplicated head, and SGF and SGF+
produce unnatural neck configurations.
Hindsight Forcing maintains consistent appearance and plausible structure as the giraffe changes its head and neck pose.
Together, these examples demonstrate coherent continuation
with meaningful motion and fewer structural artifacts.
The corresponding videos are provided in the supplementary material.

\begin{table*}[t]
\centering
\caption{
Component ablation of Hindsight Forcing to examine the contribution
of each design choice.
Starting from Causal Forcing, we progressively introduce
full-rollout history, differentiable context writing,
and the latent straight-through bridge.
Scores are scaled by 100.
Bold and underline indicate the best and second-best results, respectively. The gradient columns indicate feedback from future chunks.
}
\label{tab:component_ablation}
\vspace{-2pt}
\resizebox{\textwidth}{!}{%
\begin{tabular}{@{}l ccc ccc ccc@{}}
\toprule
\toprule
\multirow{2}{*}{\textbf{Method}}
& \multicolumn{3}{c}{\textbf{Component Configuration}}
& \multicolumn{3}{c}{\textbf{Overall Performance}}
& \multicolumn{3}{c}{\textbf{Specific Dimensions}} \\
\cmidrule(lr){2-4} \cmidrule(lr){5-7} \cmidrule(lr){8-10}
& \shortstack{\textbf{Full-Rollout}\\\textbf{History}}
& \shortstack{\textbf{Gradients to}\\\textbf{Writer}}
& \shortstack{\textbf{Gradients to}\\\textbf{Denoiser}}
& \shortstack{\textbf{VBench}\\\textbf{Total} $\uparrow$}
& \shortstack{\textbf{Quality}\\\textbf{Score} $\uparrow$}
& \shortstack{\textbf{Semantic}\\\textbf{Score} $\uparrow$}
& \shortstack{\textbf{Dynamic}\\\textbf{Degree} $\uparrow$}
& \shortstack{\textbf{Overall}\\\textbf{Consistency} $\uparrow$}
& \shortstack{\textbf{Temporal}\\\textbf{Flickering} $\uparrow$} \\
\midrule
Causal Forcing~\cite{zhu2026causal}
& $\times$ & $\times$ & $\times$
& 84.08 & 84.81 & \textbf{81.15}
& 84.17 & 26.64 & 98.05 \\

Full rollout + detach
& \checkmark & $\times$ & $\times$
& \underline{85.03} & \underline{86.05} & 80.96
& \underline{84.67} & \underline{26.73} & \underline{98.19} \\

Writer-only
& \checkmark & \checkmark & $\times$
& 84.04 & 85.23 & 79.30
& 73.33 & 26.53 & 98.03 \\
\midrule
\textbf{Hindsight Forcing (Ours)}
& \textbf{\checkmark} & \textbf{\checkmark} & \textbf{\checkmark}
& \textbf{85.74} & \textbf{86.89} & \underline{81.11}
& \textbf{92.22} & \textbf{26.92} & \textbf{99.25} \\
\bottomrule
\bottomrule
\end{tabular}%
}
\end{table*}

\begin{figure*}[t]
    \centering
    \includegraphics[width=\textwidth]{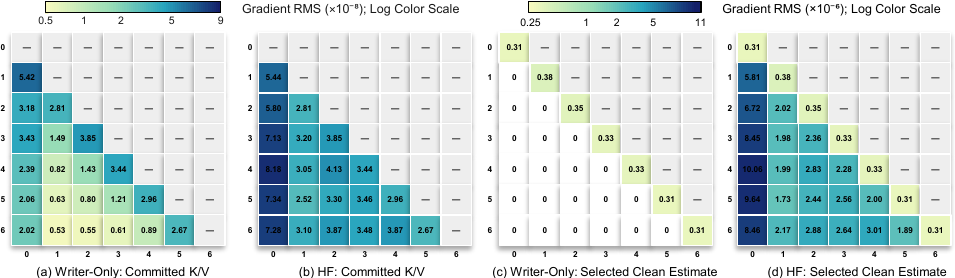}
    \caption{
        Cross-chunk gradient propagation under  Writer-only and Hindsight Forcing (HF) with gradient RMS measured on committed K/V (a, b) and selected clean estimates (c, d).
        Rows represent loss chunks, while columns represent source chunks
        receiving gradients. Colors indicate gradient magnitude on logarithmic
        scales, with different units for committed K/V ($\times 10^{-8}$)
        and selected clean estimates ($\times 10^{-6}$).
        Compared with Writer-only, Hindsight Forcing enables stronger and broader
        cross-chunk gradient propagation.
    }
    \label{fig:gradient_heatmap}
\end{figure*}

\subsection{Quantitative Results}
\label{sec:quantitative}

\subsubsection{Comparison with Existing Methods}
\label{sec:comparison}

As shown in \Cref{tab:main_results}, Hindsight Forcing achieves
the highest VBench Total and Quality scores of 85.74 and 86.89,
exceeding Reward Forcing by 0.67 and 0.61 points, respectively.
Its Semantic Score of 81.11 remains close to the highest score of 81.15, showing that improved visual quality preserves strong prompt alignment.

Limited motion can simplify future continuation and reduce
opportunities for errors to accumulate, but such conservative
generation restricts video dynamics.
SGF and SGF+ obtain Dynamic Degrees of 74.72 and 55.00,
respectively, whereas Hindsight Forcing reaches 92.22.
Notably, duplicating the backbone in SGF+ does not ensure
dynamic continuation: Hindsight Forcing surpasses it by 37.22 points
in Dynamic Degree and 1.99 points in VBench Total,
without additional student parameters.
RAVEN achieves a comparable Dynamic Degree of 91.67,
but its Quality Score of 84.71 trails Hindsight Forcing by 2.18 points.
These results highlight the importance of improving visual quality
while sustaining motion, supporting our approach of extending
future supervision to history generation for dynamic and coherent
continuation.

\subsubsection{Ablation and Mechanism Analysis}
\label{sec:ablation}

To understand the sources of these gains, we disentangle
the contributions of history construction, context writing,
and future supervision of historical denoising.

\noindent\textbf{Component Ablation.}
As shown in \Cref{tab:component_ablation}, full-rollout history
with a detached cache achieves a VBench Total of 85.03,
compared with 84.08 for Causal Forcing.
However, enabling differentiable context writing alone
reduces the score to 84.04, indicating that restoring
history-encoding gradients does not necessarily improve generation.
Adding the latent straight-through bridge raises the Total
to 85.74 and Dynamic Degree from 73.33 to 92.22.
This comparison preserves the full-rollout history construction
and differentiable writer, isolating the benefit of extending
future supervision to historical denoising.
The improvements in both overall performance and motion
support jointly optimizing how history is generated and encoded.

\noindent\textbf{Cross-Chunk Gradient Propagation.}
To examine the underlying mechanism, we compare Writer-only
and Hindsight Forcing using identical model weights and inputs,
changing only the gradient paths.
As illustrated in \Cref{fig:gradient_heatmap}, both configurations
allow future losses to reach historical K/V states.
For Writer-only, gradients at the selected clean estimates
remain confined to the diagonal, corresponding to direct
supervision of the current chunk.
Hindsight Forcing additionally produces nonzero gradients for all 21
future-to-history chunk pairs in the seven-chunk sequence,
including dependencies across nonadjacent chunks.
This verifies that the bridge extends downstream feedback
beyond context encoding to the denoising computations
that produced the historical content.

\noindent\textbf{Gradient Scale.}
\Cref{tab:alpha_ablation} examines the strength of this feedback.
The default $\alpha=1$ achieves the highest Total and Quality
scores among the tested settings.
Reducing $\alpha$ to 0.25 or 0.5 lowers
Dynamic Degree, while increasing it to 1.5 also reduces
both overall performance and motion.
These results show that stronger history gradients do not
automatically yield better generation and favor the unscaled
straight-through gradient used by our default configuration.

\subsubsection{Computational Efficiency}
\begin{figure}[t]
    \centering
    \includegraphics[width=0.9\linewidth]{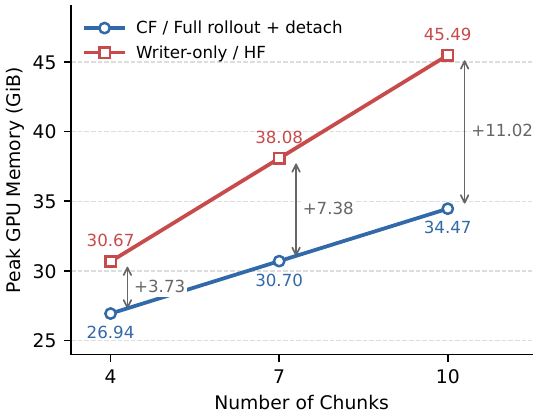}
    \caption{Peak training memory versus sequence length.
    We measure per-GPU peak allocated memory under matched training
    settings with 4, 7, and 10 chunks. Causal Forcing and Full rollout + detach exhibit nearly
identical memory usage across all tested sequence lengths,
as do Writer-only and Hindsight Forcing.
    % CF and Full rollout + detach have identical peaks;
    % Writer-only and HF overlap at the displayed precision.
    % Memory grows approximately linearly over the tested lengths.
    % HF extends future feedback to historical denoising with
    % negligible additional peak memory over Writer-only.
    % Arrows indicate the memory difference between HF and CF in GiB.
    }
    \label{fig:memory_scaling}
\end{figure}

As reported in \Cref{tab:main_results}, Hindsight Forcing requires no additional
student parameters and uses 38.08\,GiB of peak GPU memory
for seven-chunk training, compared with 30.70\,GiB for Causal Forcing.
Across the tested sequence lengths, its memory usage grows
approximately linearly and closely matches Writer-only
(\Cref{fig:memory_scaling}), indicating little additional memory
cost for extending supervision to historical denoising.
These gradient paths are established within the rollout
without separate trajectory replay.
At inference, Hindsight Forcing generates a video in 5.619 seconds,
comparable to Causal Forcing's 5.603 seconds,
while preserving the standard autoregressive generation
procedure without additional computation.

\begin{table}[t]
\centering
\caption{
Effect of the history gradient scale $\alpha$.
All variants use EMA checkpoints after 150 generator updates and the same evaluation settings.
Scores are scaled by 100.
Bold and underline indicate the best and second-best results, respectively.
}
\label{tab:alpha_ablation}
\vspace{-2pt}
\resizebox{0.92\columnwidth}{!}{%
\begin{tabular}{@{}c cccc@{}} % 第一列改为居中对齐 'c'
\toprule
\toprule
\multirow{2}{*}{\textbf{$\alpha$}}
& \textbf{VBench} & \textbf{Quality} & \textbf{Dynamic} & \textbf{Overall} \\
& \textbf{Total} $\uparrow$ & \textbf{Score} $\uparrow$ & \textbf{Degree} $\uparrow$ & \textbf{Consistency} $\uparrow$ \\
\midrule
0.25 & 83.33 & 83.99 & 51.67 & 26.58 \\
0.5  & 83.54 & 84.16 & 52.78 & 26.88 \\
1.5  & \underline{83.95} & \underline{84.86} & \underline{68.89} & \textbf{26.93} \\
\midrule
\textbf{1.0 (default)} & \textbf{85.74} & \textbf{86.89} & \textbf{92.22} & \underline{26.92} \\
\bottomrule
\bottomrule
\end{tabular}%
}
\end{table}

\section{Limitations and Further Discussion}
\label{sec:discussion}

Despite these promising results, several aspects of Hindsight Forcing remain open for further exploration.
First, our analysis focuses on a fixed gradient scale $\alpha$
shared across chunks and denoising steps.
The sensitivity observed in \Cref{tab:alpha_ablation}
motivates exploring whether adapting this scale to different
chunks or denoising steps can improve credit assignment.
Second, although retaining only one denoising graph per chunk
substantially reduces activation storage compared with full BPTT,
the number of retained denoising and context-writing graphs
still grows linearly with the number of rollout chunks.

However, we argue that effective training for streaming generation does not inherently require long rollouts.
Latency-sensitive streaming applications favor a bounded KV budget, limiting the historical context available when generating each new chunk~\cite{yang2026longlive}.
This motivates training on short rollouts that cover
history construction within the cache budget and the
continuation conditioned on that history.
Such rollouts provide downstream supervision to jointly
optimize context writing and historical denoising.
With sufficient data diversity, this offers a practical
strategy for learning to generate useful history without
matching training samples to the full deployment duration.
An important direction is therefore to investigate how
short-rollout training under bounded historical context
generalizes to sustained streaming generation.

\section{Conclusion}
\label{sec:conclusion}

We have presented Hindsight Forcing, a framework that extends future supervision from history encoding to history generation in autoregressive video diffusion.
Through a latent straight-through bridge, Hindsight Forcing preserves full-rollout history in the forward pass while routing downstream gradients to selected historical denoising steps. This enables joint optimization of history generation and
context writing without retaining the full denoising trajectory graph, introducing additional model parameters, or requiring
separate trajectory replay. Experiments demonstrate improved visual quality and temporal dynamics without altering the standard streaming inference procedure, while component ablations validate the benefit of supervising historical denoising.
More importantly, Hindsight Forcing offers a design insight for future autoregressive video diffusion frameworks: generated history should be optimized for its ability to support coherent future continuation.

\section*{LLM Usage Statement}
In accordance with the CVPR 2027 LLM policy, we disclose the use of LLMs and AI tools in preparing this submission, for which the authors take full responsibility. 

\begin{itemize}
    \item \textbf{Writing, Related Work, and Articulation of Ideas:} Google Gemini was used in an assistive capacity to edit and refine the language, clarity, and phrasing of author-written text across the manuscript and supplementary materials. All core research ideas, mathematical formulations, literature selection, citations, and conclusions were independently conceived and verified by the authors.
    \item \textbf{Code and Implementation:} OpenAI Codex (powered by GPT-6 Astra) was used in an agentic capacity under direct author supervision to perform multi-step codebase implementation, module refactoring, and debugging tasks. All resulting code was thoroughly reviewed, edited, tested, and verified by the authors.
    \item \textbf{Experiments and Figures:} All experimental designs, training and evaluation runs on GPUs, and result analyses were conducted by the authors. OpenAI Codex (GPT-6 Astra) and Google Gemini were used in an assistive capacity to refine scripts for author-designed experiments, plotting scripts, diagram labels, and figure captions. All qualitative video frames are direct outputs of the evaluated models, with no generative visual editing applied.
    \item \textbf{Data:} No LLMs or AI tools were used for dataset creation, synthetic data generation, or data annotation.
\end{itemize}

{
    \small
    \bibliographystyle{ieeenat_fullname}
    \bibliography{main}
}

\clearpage
\setcounter{page}{1}
\maketitlesupplementary

\begin{table*}[t]
\centering
\small
\caption{
Inference-time long-horizon extrapolation configurations used for one-minute generation (80 chunks, 240 latent frames). Budget denotes the active cache window in latent frames (including the current 3-latent-frame chunk).
}
\label{tab:long_kv_config}
\begin{tabular}{@{}c c c@{}}
\toprule
\toprule
Method & Budget (Latents) & Long-Horizon Inference Strategy \\
\midrule
CausVid~\cite{yin2025slow} & 21 & Native overlap extension (3-latent-frame overlap) \\
Self Forcing~\cite{huang2025self} & 21 & Sink+FIFO (RoPE span 21) \\
Causal Forcing~\cite{zhu2026causal} & 21 & Sink+FIFO (RoPE span 21) \\
Reward Forcing~\cite{lu2026reward} & 9 & Native EMA-Sink ($\text{decay}=0.999$) \\
RAVEN~\cite{lu2026raven} & 21 & Sink+FIFO (RoPE span 21) \\
SGF~\cite{zhuang2026self} & 12 & Sink+FIFO (RoPE span 21) \\
SGF+~\cite{su2026sgf+} & 12 & Sink+FIFO (RoPE span 12) \\
\midrule
Hindsight Forcing (Ours) & 12 & Sink+FIFO (RoPE span 12) \\
Hindsight Forcing Long (Ours) & 12 & Sink+FIFO (RoPE span 12) \\
\bottomrule
\bottomrule
\end{tabular}
\end{table*}

\begin{figure*}[!t]
    \centering
    \includegraphics[width=\textwidth]{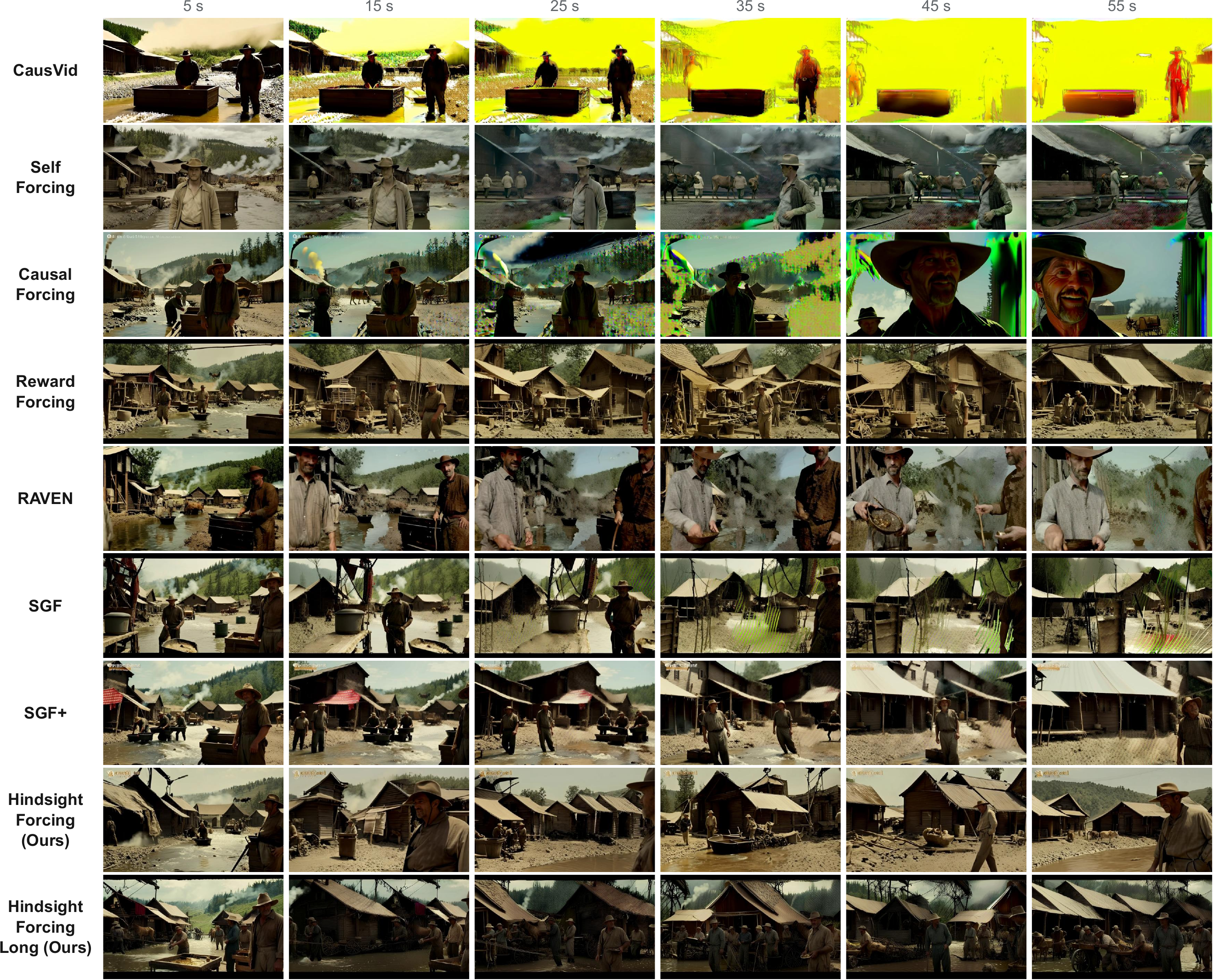}
    \caption{
       Qualitative comparison on one-minute video generation for the gold-rush camp scene. Frames sampled at 5, 15, 25, 35, 45, and 55 seconds show that CausVid, Self Forcing, Causal Forcing, and SGF suffer from color drift and color collapse, Reward Forcing degrades in quality at later timestamps, and RAVEN and SGF+ develop artifacts and quality degradation over time. In contrast, both Hindsight Forcing and Hindsight Forcing Long preserve a coherent camp layout, consistent lighting, and sustained motion across the entire minute.
    }
    \label{fig:qualitative_long_gold_rush}
\end{figure*}

\begin{figure*}[!t]
    \centering
    \includegraphics[width=\textwidth]{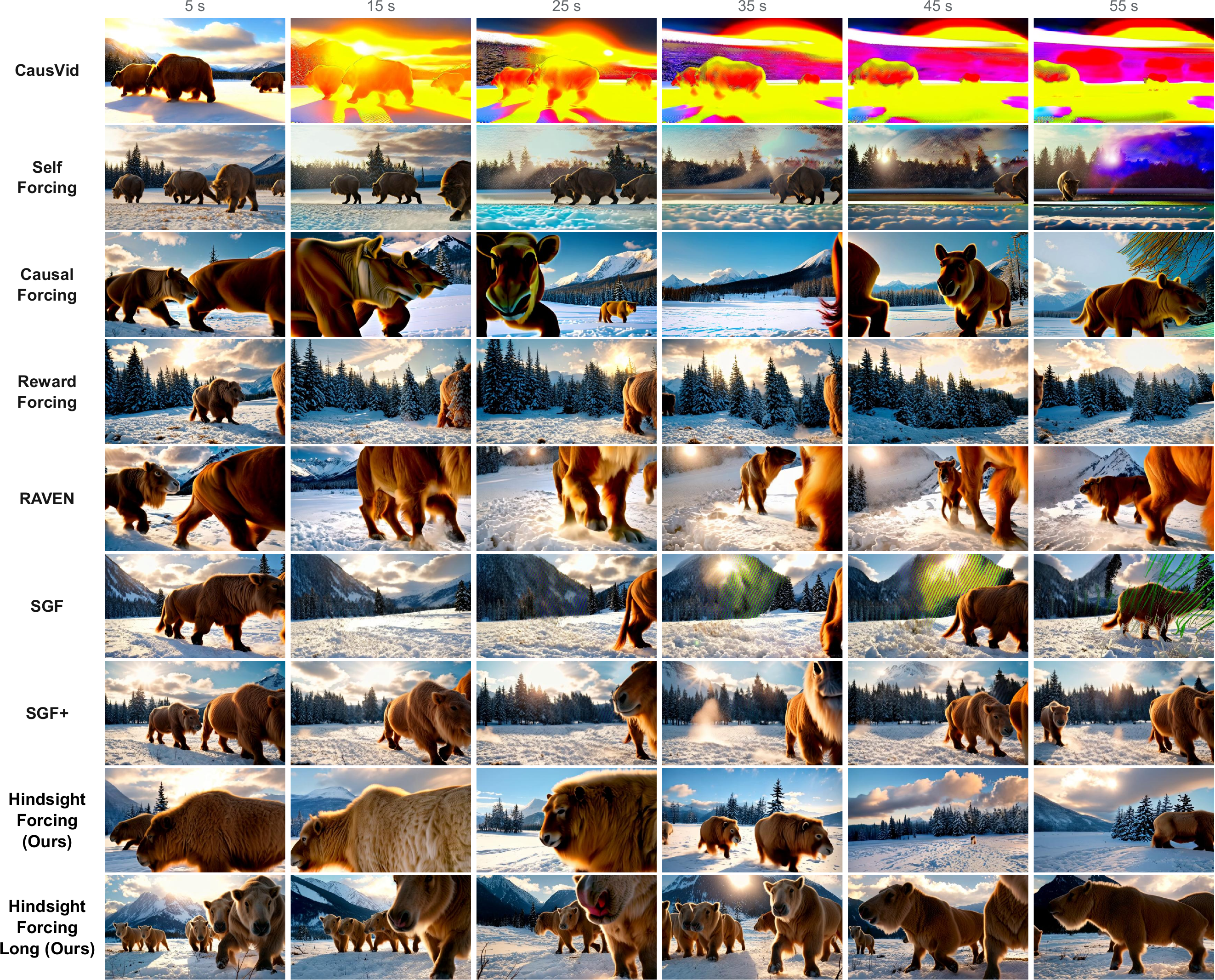}
    \caption{
       Qualitative comparison on one-minute video generation for the snowy mammoth scene. Frames sampled at 5, 15, 25, 35, 45, and 55 seconds show that CausVid, Self Forcing, Causal Forcing, and SGF suffer from severe color drift, Reward Forcing exhibits abrupt animal appearances, and RAVEN and SGF+ experience noticeable texture drift. In contrast, both Hindsight Forcing and Hindsight Forcing Long preserve natural coloration, clean snow and fur details, and sustained motion across the entire minute.
    }
    \label{fig:qualitative_long_mammoths}
\end{figure*}

\begin{table*}[t]
\centering
\caption{
Quantitative comparison on one-minute video generation.
All scores are scaled by 100.
}
\label{tab:long_video_results}
\vspace{-2pt}
\resizebox{\textwidth}{!}{%
\begin{tabular}{@{}l cccccc@{}}
\toprule
\toprule
Method
& \shortstack{Subject\\Consistency $\uparrow$}
& \shortstack{Motion\\Smoothness $\uparrow$}
& \shortstack{Imaging\\Quality $\uparrow$}
& \shortstack{Aesthetic\\Quality $\uparrow$}
& \shortstack{Dynamic\\Degree $\uparrow$}
& \shortstack{Background\\Consistency $\uparrow$} \\
\midrule
CausVid~\cite{yin2025slow}
& 96.08 & 97.76 & 59.23 & 49.58 & 44.22 & 96.55 \\

Self Forcing~\cite{huang2025self}
& 96.50 & 98.03 & 67.37 & 57.83 & 52.45 & 95.86 \\

Causal Forcing~\cite{zhu2026causal}
& 95.15 & 97.06 & 68.06 & 55.24 & 79.71 & 94.51 \\

Reward Forcing~\cite{lu2026reward}
& 95.14 & 97.41 & 64.73 & 56.40 & 90.76 & 94.65 \\

RAVEN~\cite{lu2026raven}
& 92.79 & 96.30 & 66.77 & 55.60 & 86.20 & 93.67 \\

SGF~\cite{zhuang2026self}
& 96.16 & 98.18 & 69.01 & 57.54 & 60.83 & 95.54 \\

SGF+~\cite{su2026sgf+}
& 97.01 & 98.47 & 68.86 & 58.35 & 45.18 & 95.94 \\

\midrule
Hindsight Forcing (Ours)
& 93.86 & 96.95 & 61.92 & 55.38 & 89.06 & 94.19 \\

Hindsight Forcing Long (Ours)
& 96.34 & 98.44 & 69.67 & 57.74 & 75.00 & 95.01 \\
\bottomrule
\bottomrule
\end{tabular}%
}
\end{table*}

\section{Long-Horizon Video Generation}
\label{sec:supp_long_video}

\subsection{Evaluation Setup and Model Variants}

\noindent\textbf{One-Minute Evaluation Protocol.}
We evaluate long-horizon extrapolation on the first 128 prompts (in release order) from MovieGenVideoBench~\cite{polyak2024movie}, generating one video per prompt (128 videos in total) from the released expanded prompts. All methods share identical evaluation prompts and matched random seeds (generated deterministically from a fixed base seed, prompt index, and sample index) for paired comparison. All methods generate videos at a resolution of $832\times480$ and 16\,FPS, producing 957 decoded frames (59.81\,s, or approximately one minute). Streaming KV-cache methods generate 240 latent frames in a continuous rollout of 80 chunks, with three latent frames and four denoising steps per chunk.
CausVid~\cite{yin2025slow} instead uses its native
overlap-extension procedure to produce the same decoded
video length.

We score all generated videos using the official VBench-Long
evaluator~\cite{huang2024vbench,huang2025vbenchpp}
(\texttt{vbench2\_beta\_long}, VBench package v0.1.5)
across six dimensions: Subject Consistency, Background Consistency, Motion Smoothness, Dynamic Degree, Imaging Quality, and Aesthetic Quality. Following this evaluation protocol without static filtering or scene-based splitting, each 957-frame video is partitioned into 30 two-second clips of 32 frames at 16\,FPS (with the final clip taking the last 32 frames and overlapping the preceding clip by 3 frames). Clip-level scores are averaged per video and then across all 128 videos. Note that Subject Consistency and Background Consistency use the within-clip component of \texttt{vbench2\_beta\_long}, and thus reflect local temporal consistency over the one-minute rollout.

\medskip
\noindent\textbf{Model Variants: Hindsight Forcing vs.\
Hindsight Forcing Long.}
Both variants share the same backbone, initialization,
training prompts, global batch size, gradient scale
$\alpha=1$, and long-horizon inference configuration.
We evaluate their EMA checkpoints after 150 generator updates.
In the main paper, Hindsight Forcing is built directly on the
Causal Forcing~\cite{zhu2026causal} codebase and training
setup without introducing extra architectural modifications
or auxiliary training tricks.
This controlled design isolates and verifies the
effectiveness of the latent straight-through bridge while
keeping the evaluation protocol directly aligned with
existing baselines such as Causal Forcing and
RAVEN~\cite{lu2026raven}, whose primary VBench evaluations
focus on five-second (81-frame) generation.
Specifically, a selected denoising step $s\in\{1,\ldots,K\}$
is shared across chunks and ranks, each chunk executes the
full $K$-step rollout, and the final output $x_j$ provides
forward history while downstream gradients are routed to
$z_j^{(s)}$ through the bridge.

While this formulation achieves the strongest VBench Total,
Quality, and Dynamic Degree scores in our five-second
comparison, one-minute video generation extends the
seven-chunk training horizon to 80 chunks under a bounded
four-chunk KV cache.
To achieve strong long-horizon performance without altering
the core Hindsight Forcing framework, we introduce Hindsight
Forcing Long by incorporating a single implementation trick
from the codebases of the leading long-horizon baselines
SGF~\cite{zhuang2026self} and SGF+~\cite{su2026sgf+}, namely
\emph{per-rank independent step sampling}.
Under this modification, each rank independently samples
$s\in\{1,\ldots,K\}$, shared across chunks within its rollout,
rather than synchronizing one denoising step across all ranks. This exposes each generator update to supervision across multiple refinement stages within a batch.

All remaining training and inference components of Hindsight
Forcing are preserved without modification.
Each chunk still executes the full $K$-step rollout, retains
the computation graph only at the selected denoising call
$z_j^{(s)}=G_\theta(\mathrm{sg}(u_j^{(s)}),\sigma_s,c,H_{<j})$
with all other steps $r\neq s$ detached, and applies the
latent straight-through bridge
$h_j=\tilde{x}_j=\mathrm{sg}(x_j)+[z_j^{(s)}-\mathrm{sg}(z_j^{(s)})]$
($\alpha=1$) followed by the context-noise operation and
differentiable context writing in \Cref{eq:context_writing}.
Unlike SGF and SGF+, which adopt selected-step history
construction and detach historical latents before context
encoding, Hindsight Forcing Long retains full-rollout
forward history and preserves the gradient path
$\mathcal{L}_{>j}\rightarrow C_j\rightarrow \tilde{x}_j\rightarrow z_j^{(s)}$
back to the historical denoiser parameters~$\theta$ without
a separate replay pass or an additional writer network.

\medskip
\noindent\textbf{Long-Horizon Inference KV-Cache Configurations.}
Because retaining all 80 chunks (240 latent frames) in the KV cache is unsuitable for streaming extrapolation, all methods use bounded caches or segmented extension at inference, summarized in \Cref{tab:long_kv_config}. For the Sink+FIFO variants, the cache budget includes the current chunk: a 12-latent-frame (4-chunk) budget retains the initial sink chunk, the 2 most recent historical chunks, and the current chunk; a 21-latent-frame (7-chunk) budget retains the initial sink chunk, the 5 most recent historical chunks, and the current chunk. For Self Forcing~\cite{huang2025self}, Causal Forcing~\cite{zhu2026causal}, and RAVEN~\cite{lu2026raven}, we evaluate their respective checkpoints using a 7-chunk (21-latent-frame) Sink+FIFO rolling-cache path adapted from the SGF~\cite{zhuang2026self} implementation (RoPE span of 21 latent frames). SGF~\cite{zhuang2026self} uses a 4-chunk (12-latent-frame)
Sink+FIFO cache with a RoPE span of 21 latent frames, whereas
SGF+~\cite{su2026sgf+}, Hindsight Forcing, and Hindsight
Forcing Long use a 4-chunk (12-latent-frame) Sink+FIFO cache
with a RoPE span of 12 latent frames. CausVid~\cite{yin2025slow} and Reward Forcing~\cite{lu2026reward} retain their native overlap-extension (21-latent-frame segments passing 3 boundary latent frames) and EMA-Sink (9-latent-frame budget, $\text{decay}=0.999$) strategies, respectively.

\subsection{Qualitative Comparisons}

\Cref{fig:qualitative_long_gold_rush,fig:qualitative_long_mammoths} present qualitative comparisons across one-minute rollouts sampled from 5\,s to 55\,s on two prompts.

\noindent\textbf{Gold-Rush Camp (\Cref{fig:qualitative_long_gold_rush}).}
Clear differences emerge among the baselines as generation extends far beyond the 5-second training horizon. CausVid~\cite{yin2025slow}, Self Forcing~\cite{huang2025self}, Causal Forcing~\cite{zhu2026causal}, and SGF~\cite{zhuang2026self} all experience progressive color drift and color collapse: CausVid turns bright yellow and overexposed from 15\,s onward, Self Forcing accumulates dark greenish-magenta tint and noise after 35\,s, Causal Forcing develops cyan and green artifacts before collapsing at 25\,s, and SGF shifts into darkened orange-red contrast collapse. Reward Forcing~\cite{lu2026reward} keeps relatively stable coloration but degrades noticeably in visual sharpness and texture quality at later timestamps (35--55\,s). RAVEN~\cite{lu2026raven} and SGF+~\cite{su2026sgf+} preserve early scene structure, yet accumulate visible artifacts and quality degradation over time. In contrast, both Hindsight Forcing and Hindsight Forcing Long avoid color collapse and late-stage degradation, maintaining a coherent camp layout, consistent lighting, and sustained prospector motion, with Hindsight Forcing Long achieving even cleaner late-stage details.

\noindent\textbf{Snowy Mammoths (\Cref{fig:qualitative_long_mammoths}).}
Similar failure patterns appear on the snowy mammoth scene. CausVid, Self Forcing, Causal Forcing, and SGF again suffer from severe color drift over time, including neon yellow-magenta saturation in CausVid, blue-magenta sky tinting in Self Forcing, unnatural color shifts and structural distortion in Causal Forcing, and prominent green-yellow artifacts in SGF. Reward Forcing exhibits abrupt animal appearances across timestamps. Meanwhile, RAVEN and SGF+ accumulate noticeable texture drift across the snow, forest, and fur regions at later timestamps. In contrast, both Hindsight Forcing and Hindsight Forcing Long avoid color drift and texture drift, preserving natural coloration, clean snow and fur details, and sustained motion across the entire minute.

\subsection{Quantitative Results}

\Cref{tab:long_video_results} reports the one-minute quantitative comparison across all 128 prompts on MovieGenVideoBench~\cite{polyak2024movie}, quantitatively reflecting the fidelity, consistency, and motion trade-offs seen in \Cref{fig:qualitative_long_gold_rush,fig:qualitative_long_mammoths}.

\medskip
\noindent\textbf{Consistency and Motion Trade-offs on One-Minute Rollouts.}
Under bounded-cache extrapolation, existing baselines struggle to preserve visual fidelity and dynamic motion together. Several methods with strong local consistency or aesthetic scores suffer substantial motion decay over one minute: CausVid~\cite{yin2025slow}, Self Forcing~\cite{huang2025self}, and SGF+~\cite{su2026sgf+} drop to Dynamic Degrees of 44.22, 52.45, and 45.18, respectively, while SGF~\cite{zhuang2026self} reaches 60.83. Conversely, baselines with higher Dynamic Degrees such as Reward Forcing~\cite{lu2026reward} (90.76) and RAVEN~\cite{lu2026raven} (86.20) exhibit lower Imaging Quality (64.73 and 66.77) and weaker consistency (for example, RAVEN drops to 92.79 Subject Consistency and 96.30 Motion Smoothness). Note also that because Subject Consistency and Background Consistency evaluate the within-clip component over 2-second windows, locally smooth or color-collapsed videos can still retain high local consistency scores, as illustrated by CausVid (96.55 Background Consistency despite 59.23 Imaging Quality and 49.58 Aesthetic Quality).

\medskip
\noindent\textbf{Hindsight Forcing vs.\ Hindsight Forcing Long.}
Evaluating the standard 5-second Hindsight Forcing model at one minute preserves strong dynamics (89.06 Dynamic Degree), yet across the full 128-prompt benchmark its average Imaging Quality (61.92) and Subject Consistency (93.86) degrade under $12\times$ rollout extrapolation and four-chunk cache eviction. In contrast, Hindsight Forcing Long trades a moderate reduction in Dynamic Degree (75.00 vs.\ 89.06, while still outperforming SGF+ by +29.82 points and SGF by +14.17 points) for substantial gains across every visual quality and consistency metric. It raises Imaging Quality to 69.67 (+7.75, the highest among all methods), Subject Consistency to 96.34 (+2.48), Aesthetic Quality to 57.74 (+2.36), Motion Smoothness to 98.44 (+1.49, nearly matching 98.47 for SGF+), and Background Consistency to 95.01 (+0.82), resulting in far better overall one-minute performance. Compared with the Causal Forcing initialization~\cite{zhu2026causal}, Hindsight Forcing Long also improves all five consistency and visual quality metrics (for example, +1.61 in Imaging Quality and +2.50 in Aesthetic Quality) while avoiding its severe color drift and structural distortion.

\noindent\textbf{Supervising Historical Denoising Under Identical Extrapolation Settings.}
A direct comparison between SGF+~\cite{su2026sgf+} and
Hindsight Forcing Long is particularly informative because
both share the same four-chunk (12-latent-frame) Sink+FIFO
cache with a RoPE span of 12 latent frames and per-rank
independent step sampling.
Using selected-step history with detached historical latents
in SGF+, even with two-pass replay and +100\% extra
parameters, causes Dynamic Degree to fall to 45.18.
In contrast, preserving full-rollout history and keeping
future supervision connected to the historical denoiser via
the latent straight-through bridge in Hindsight Forcing Long
improves Imaging Quality from 68.86 to 69.67 while boosting
Dynamic Degree by nearly 30 points (75.00 vs.\ 45.18) with
zero additional parameters, confirming that jointly preserving
full-rollout history and supervising history generation
remains effective under long-horizon rolling caches.

\begin{algorithm}[t]
\caption{Hindsight Forcing: One Student Update}
\label{alg:hindsight_forcing}
\small
\begin{algorithmic}[1]
\Require Denoiser $G_\theta$ and context writer $W_\theta$
         sharing parameters $\theta$
\Require Frozen teacher $D_T$, critic $D_\phi$, condition $c$
\Require $N$ chunks, $K$ denoising steps, bridge scale $\alpha$

\State Sample a denoising step $s \in \{1,\ldots,K\}$
       shared across chunks
\State Initialize $H_{<0} \gets \varnothing$

\For{$j = 0,\ldots,N-1$}
    \State Run the full $K$-step denoising rollout
           conditioned on $c$ and $H_{<j}$
    \State Obtain the final output $x_j = \hat{x}_j^{(K)}$
           and selected estimate $z_j^{(s)} = \hat{x}_j^{(s)}$
    \State Retain only the selected denoising call's graph,
           with noisy latent inputs detached
    \State $\tilde{x}_j \gets \operatorname{sg}(x_j)
           + \alpha\bigl[z_j^{(s)}
           - \operatorname{sg}(z_j^{(s)})\bigr]$
    \State Construct $C_j$ using \Cref{eq:context_writing}
           with $h_j = \tilde{x}_j$, retaining its graph
    \State $H_{<j+1} \gets H_{<j} \cup \{C_j\}$
\EndFor

\State $Z \gets
       \operatorname{Concat}(z_0^{(s)},\ldots,z_{N-1}^{(s)})$
\State Evaluate $D_T$ and $D_\phi$ on the noisy
       $\operatorname{sg}(Z)$ to obtain
       $\mu_{\mathrm{real}}$ and $\mu_{\mathrm{fake}}$
\State $g \gets \operatorname{sg}\!\left[
       \operatorname{Normalize}
       (\mu_{\mathrm{fake}}-\mu_{\mathrm{real}})
       \right]$
\State $\mathcal{L}_{\mathrm{student}} \gets
       \dfrac{1}{2D}
       \left\|Z-\operatorname{sg}(Z-g)\right\|_2^2$
\State Update $\theta$ using
       $\nabla_\theta\mathcal{L}_{\mathrm{student}}$,
       keeping $D_T$ and $D_\phi$ fixed
\State Update the generator EMA
\end{algorithmic}
\end{algorithm}

\begin{table*}[t]
\centering
\small
\caption{
Matched training profiling on NVIDIA H100 80\,GB GPUs (7 chunks, global batch size 64). Wall-clock times cover full student and critic updates with CPU FP32 EMA enabled.
}
\label{tab:training_profiling}
\begin{tabular}{@{}l ccc@{}}
\toprule
\toprule
Method & Student Update (s) & Critic Update (s) & Peak Memory (GiB) \\
\midrule
Causal Forcing (matched) & 64.80 & 28.47 & 30.70 \\
Full rollout + detach & 69.73 & 33.46 & 30.70 \\
Writer-only & 79.08 & 33.67 & 38.08 \\
Hindsight Forcing (Ours) & 77.96 & 33.63 & 38.08 \\
\bottomrule
\bottomrule
\end{tabular}
\end{table*}

\section{Algorithm and Implementation Details}
\label{sec:supp_implementation}

\subsection{Training Algorithm}

\Cref{alg:hindsight_forcing} summarizes one student update of
Hindsight Forcing. Within each training rollout, a single
denoising step $s\in\{1,\ldots,K\}$ is sampled and shared
across all $N$ temporal chunks of the sequence. During the
sequential forward pass for chunk~$j$, the generator executes
the full $K$-step rollout conditioned on~$c$ and the
accumulated historical cache~$H_{<j}$. To retain only $N$
denoising computation graphs rather than $NK$, steps $r\neq s$
are executed without gradient tracking, and stop-gradient is
applied to the noisy latent input $u_j^{(s)}$, keeping only the
computation graph of the selected denoising call
$z_j^{(s)}=G_\theta(\operatorname{sg}(u_j^{(s)}),\sigma_s,c,H_{<j})$.

Crucially, because standard DMD supervision already requires retaining the selected-step graph for $Z=[z_0^{(s)},\ldots,z_{N-1}^{(s)}]$, reusing the same step~$s$ for the latent straight-through bridge introduces no extra denoising computation graphs compared with detached baselines. In the forward pass, $\tilde{x}_j$ equals the full-rollout output $x_j$ so that subsequent chunks always condition on full-rollout history encoded via \Cref{eq:context_writing}. In the backward pass, retaining the graph of $W_\theta$ and its connection through $\tilde{x}_j$ enables downstream chunk losses $\mathcal{L}_{>j}$ to supervise both context encoding and historical denoising on shared parameters~$\theta$. At inference, the bridge is inactive and no computation graphs are retained, leaving the standard $K$-step streaming KV-cache rollout unchanged.

\subsection{Training and Evaluation Configurations}

\noindent\textbf{Backbone and Optimization.}
Unless otherwise noted, all trained models follow the Causal Forcing~\cite{zhu2026causal} setup on Wan2.1-T2V-1.3B~\cite{wan2025wan} with a frozen Wan2.1-T2V-14B teacher and a trainable Wan2.1-T2V-1.3B critic, using 7-chunk sequences (3 latent frames and $K=4$ denoising steps per chunk), global batch size~64, 5 critic updates per generator update, and $\alpha=1$ for 150 generator updates. Both the generator and critic are trained in \texttt{bfloat16} precision using AdamW ($\beta_1=0.0$, $\beta_2=0.999$, weight decay~$0.01$) with gradient clipping at~$10.0$, learning rates of $2.0\times10^{-6}$ (generator) and $4.0\times10^{-7}$ (critic), and student EMA ($\text{decay}=0.99$) enabled after 40 generator updates.

\medskip
\noindent\textbf{Schedules, Context Noise, and DMD Supervision.}
Using a 1000-step shifted scheduler ($\text{shift}=5$), the four-step denoising schedule $[1000, 750, 500, 250]$ maps to timesteps $[1000, 937.5, 833.3, 625]$ ($\sigma=[1, 0.9375, 0.8333, 0.625]$), where the fourth prediction gives $x_j=\hat{x}_j^{(4)}$. In \Cref{eq:context_writing}, the context writer uses a zero-timestep setting, which maps to the smallest nonzero grid noise level $\sigma_c\approx 0.005$ under the shifted scheduler. For DMD supervision on~$Z$, we uniformly sample a timestep in $\{0,\ldots,999\}$, apply $\text{shift}=5$, clamp to $[20, 980]$, and share one re-noising level across the video. The critic is trained with the same re-noising distribution using the standard flow-matching loss on $\operatorname{sg}(Z)$. The teacher applies classifier-free guidance with scale~$3$ ($\mu_{\mathrm{real}}=\mu_{\mathrm{cond}}+3(\mu_{\mathrm{cond}}-\mu_{\mathrm{negative}})$, equivalent to $w=4$), while the critic uses conditional predictions without guidance. Following baseline DMD, the per-video normalized update direction is
\begin{equation}
g = \operatorname{sg}\!\left[
\frac{\mu_{\mathrm{fake}} - \mu_{\mathrm{real}}}
{\operatorname{mean}|Z - \mu_{\mathrm{real}}|}
\right].
\end{equation}

\medskip
\noindent\textbf{Matched Baseline Reproduction.}
In \Cref{tab:main_results}, Reward Forcing~\cite{lu2026reward}, RAVEN~\cite{lu2026raven}, SGF~\cite{zhuang2026self}, and SGF+~\cite{su2026sgf+} are trained from the same Causal Forcing ODE initialization on the 248K prompt set for 150 generator updates (global batch size~64, EMA evaluation). Each baseline preserves its core design. Reward Forcing uses Re-DMD with the motion-quality (MQ) score from a frozen VideoReward model based on Qwen2-VL-2B together with EMA-Sink caching, while RAVEN uses its interleaved history-and-noise sequence without CM-GRPO under its method-specific settings (two critic updates per generator update, zero weight decay, and gradient clipping at~1000). SGF uses two-pass context-reconstruction replay, and SGF+ adds full parameter separation between denoising and context writing with two-pass replay.

\medskip
\noindent\textbf{Hardware, Inference Timing, and Runtime Profiling.}
All 5-second inference times in \Cref{tab:main_results} are measured on a single NVIDIA H100 80\,GB GPU. Training peak memory in \Cref{tab:main_results} and the sequence-length scaling in \Cref{fig:memory_scaling} (4, 7, and 10 chunks) are measured on distributed H100 80\,GB GPUs with per-GPU micro-batch size~1, global batch size~64, FSDP, and activation checkpointing (with RAVEN maintaining the same micro-batch and global batch sizes in a separate distributed run). Under the same multi-GPU H100 setup (7 chunks, evenly covering $s\in\{1,2,3,4\}$), \Cref{tab:training_profiling} reports profiling results after five warmup global steps, measuring one complete student update and one critic update (including rollout, DMD evaluation, backward pass, optimizer step, and CPU FP32 EMA). Differentiable context writing increases peak memory from 30.70\,GiB to 38.08\,GiB and student update time from 69.73\,s to 79.08\,s (+13.4\% over Full rollout + detach). Hindsight Forcing measures 38.08\,GiB and 77.96\,s (+11.8\% over Full rollout + detach), showing comparable memory and runtime costs to Writer-only in this profiling because the latent straight-through bridge reuses the selected-step denoising graphs.

\subsection{Gradient Measurement}

To isolate the effect of gradient routing in \Cref{fig:gradient_heatmap}, we compare Writer-only and Hindsight Forcing using identical model weights and stochastic inputs, changing only the bridge routing ($\alpha=0$ with differentiable context writing for Writer-only vs.\ $\alpha=1$ for Hindsight Forcing). Both configurations are evaluated on the non-EMA generator checkpoint after 150 generator updates, with no parameter updates. We run 7-chunk rollouts on a single skiing prompt across 16 independent noise seeds and fix the selected denoising step to $s=2$ (corresponding to timestep~$937.5$ and $\sigma=0.9375$). Teacher signals are computed using the frozen Wan2.1-T2V-14B model, and critic predictions use the pretrained Wan2.1-T2V-1.3B initialization.

For each seed, we first compute the detached DMD update direction $g$ over all 7 chunks. To measure how much supervision from each target chunk~$k\in\{0,\ldots,6\}$ reaches each source chunk~$j\le k$, we perform 7 target-isolated backward passes per seed using the chunk-$k$ objective
\begin{equation}
\ell_k = \frac{\langle z_k^{(s)}, g_k \rangle}{D},
\end{equation}
where $D$ is the total number of scalar elements in $Z$; since $g_k$ is detached, $\nabla_{z_k^{(s)}}\ell_k = g_k/D$ matches the direct chunk-$k$ DMD gradient in \Cref{eq:total_gradient_accumulation}. Each target pass rebuilds the computation graph under identical inputs (verified to have zero forward replay difference between the two routings) and records gradients arriving at two targets per chunk~$j$: the committed key and value states $C_j=(K_j,V_j)$ aggregated across all Transformer layers ($j < k$), and the selected-step clean estimate $z_j^{(s)}$ ($j \le k$). For each pair $(k,j)$, \Cref{fig:gradient_heatmap} reports the pooled gradient root-mean-square (RMS) across all 16 seeds:
\begin{equation}
M_{kj}
= \sqrt{
\frac{\sum_{\text{seeds}} \left\|\partial \ell_k / \partial \xi_j\right\|_2^2}
{\sum_{\text{seeds}} \operatorname{numel}(\xi_j)}
},
\end{equation}
where $\xi_j$ denotes either the concatenated layer-wise $(K_j,V_j)$ tensors or $z_j^{(s)}$, and $\operatorname{numel}(\xi_j)$ is its element count.

\section{Notation}
\label{sec:supp_notation}

\Cref{tab:notation} summarizes the mathematical notation used in the main paper, grouped into rollout and history variables, DMD objective terms, and gradient-analysis quantities.

\begin{table*}[t]
\centering
\small
\renewcommand{\arraystretch}{1.12}
\caption{Summary of notation used in the main paper.}
\label{tab:notation}
\begin{tabular}{@{}l p{0.76\textwidth}@{}}
\toprule
\toprule
\textbf{Symbol} & \textbf{Description} \\
\midrule
$N$
& Number of temporal chunks in a training sequence. \\

$K$
& Number of denoising steps in a full chunk rollout. \\

$j,\ k$
& Chunk indices, with $k>j$ indicating a chunk generated after chunk $j$. \\

$r,\ s$
& Denoising step index and randomly sampled step retained for gradient propagation, respectively. \\

$\theta,\ \phi$
& Shared generator parameters and trainable diffusion critic parameters, respectively. \\

$c$
& Conditioning information, including the text prompt and an optional reference image. \\

$G_\theta,\ W_\theta$
& Denoising and context-writing calls to the same network with shared parameters $\theta$. \\

$u_j^{(r)},\ \sigma_r$
& Noisy latent input and its noise level at denoising step $r$ of chunk $j$. \\

$\hat{x}_j^{(r)}$
& Clean latent estimate predicted at denoising step $r$ of chunk $j$. \\

$x_j$
& Final clean latent output of the full rollout,
$x_j=\hat{x}_j^{(K)}$. \\

$z_j^{(s)}$
& Clean latent estimate from the selected denoising step,
$z_j^{(s)}=\hat{x}_j^{(s)}$. \\

$h_j$
& Clean latent supplied to the context writer to construct history for chunk $j$. \\

$\tilde{x}_j$
& Bridged latent, equal to $x_j$ in the forward pass and connected to $z_j^{(s)}$ through a surrogate backward path. \\

$C_j=(K_j,V_j)$
& Key and value tensors produced by context writing for chunk $j$. \\

$H_{<j}$
& Historical KV cache available before generating chunk $j$. \\

$q_{\sigma_c},\ \sigma_c$
& Context forward-noising operation and its noise level, respectively. \\

$\eta_j$
& Gaussian noise used in context preprocessing for chunk $j$. \\

$\operatorname{sg}(\cdot)$
& Stop-gradient operator, preserving the forward value while blocking backward gradients. \\

$\bar{C}_j,\ \bar{h}_j$
& Detached KV states and detached historical latents, respectively. \\

$\alpha$
& Gradient scaling factor of the latent straight-through bridge. \\

$I$
& Identity matrix. \\

\midrule
$D_T,\ D_\phi$
& Frozen bidirectional diffusion teacher and trainable diffusion critic, respectively. \\

$Z$
& Concatenated selected-step clean estimates,
$Z=[z_0^{(s)},\ldots,z_{N-1}^{(s)}]$. \\

$\sigma,\ \epsilon$
& Sampled noise level and Gaussian noise used for DMD supervision. \\

$Y_\sigma$
& Noisy sequence supplied to the teacher and critic,
$Y_\sigma=(1-\sigma)\operatorname{sg}(Z)+\sigma\epsilon$. \\

$\mu_{\mathrm{real}},\ \mu_{\mathrm{fake}}$
& Clean latent predictions from the teacher and critic, respectively. \\

$g,\ g_j$
& Detached normalized DMD update direction and its chunk-$j$ slice, respectively. \\

$D$
& Total number of scalar elements in $Z$. \\

$\mathcal{L}_{\mathrm{student}},\ \ell_j$
& Generator surrogate loss and its chunk-$j$ term, with
$\mathcal{L}_{\mathrm{student}}=\sum_j\ell_j$. \\

$\mathcal{L}_{>j}$
& Generation loss terms associated with later chunks,
$\mathcal{L}_{>j}=\sum_{k>j}\ell_k$. \\

\midrule
$\delta_{C_j}$
& Gradient from downstream losses arriving at the historical KV states,
$\delta_{C_j}=\nabla_{C_j}\mathcal{L}_{>j}$. \\

$J_{W,\theta},\ J_{G,\theta}$
& Local parameter Jacobians of context writing and the selected denoising call, with their inputs held fixed. \\

$J_{W,h}$
& Jacobian of context writing with respect to its historical latent input,
$J_{W,h}=\partial C_j/\partial h_j$. \\

$J_{k\leftarrow j}$
& Surrogate Jacobian from $z_j^{(s)}$ to $z_k^{(s)}$ along retained history-conditioning paths, including paths through intervening chunks. \\

$\Delta_\theta^{(j)}$
& Contribution to the shared parameter gradient from downstream feedback through chunk $j$'s context-writing and selected denoising calls. \\
\bottomrule
\bottomrule
\end{tabular}
\end{table*}

\end{document}